\documentclass[11pt,a4paper]{article}

\usepackage[utf8]{inputenc}
\usepackage[margin=1in]{geometry}
\usepackage{amsmath,amssymb,amsfonts}
\usepackage{booktabs}
\usepackage{graphicx}
\usepackage{microtype}
\usepackage{hyperref}
\usepackage{url}
\usepackage{cite}
\usepackage{listings}
\usepackage{xcolor}
\usepackage{algorithm}
\usepackage{algpseudocode}
\usepackage{caption}
\usepackage{subcaption}

\hypersetup{
    colorlinks=true,
    linkcolor=blue!70!black,
    citecolor=blue!70!black,
    urlcolor=blue!70!black
}

\title{\textbf{Algorithmic Scratchpads and Curriculum Staging for Arithmetic Reasoning in Tiny Transformers}\thanks{Code, model weights, and benchmark datasets are available at \url{https://github.com/SoKasl/AIMathLab}.}}

\author{
  \textbf{Sourabh Kasliwal} \\
  \texttt{sourabh.kasliwal@gmail.com}
}

\date{}

\begin{document}

\maketitle

\begin{abstract}
Autoregressive Large Language Models (LLMs) frequently struggle with deterministic multi-step algorithmic tasks such as multi-digit multiplication and long division. In this paper, we investigate the mechanics of multi-step arithmetic in compact ``Tiny'' Transformers ($\sim$10.6M non-embedding parameters, 49.3M total) trained on synthetic data across four basic operations ($+$, $-$, $\times$, $/$) unrolled as step-by-step scratchpads.

First, we establish the necessary training foundations: (1) dataloader sequence padding creates an 83\% gradient starvation artifact that collapses accuracy from 40\% to 1\%, remediated via continuous sequence packing; (2) linguistic pretraining is an essential prerequisite ($\le 2.0\%$ without it); and (3) modern architectural primitives (RoPE, RMSNorm, SwiGLU) and Sparse Mixture of Experts (MoE) substantially improve additive reasoning over baseline GPT-2. 

Second, we demonstrate that algorithmic scratchpad formulation directly dictates success. Introducing a deterministic \textit{Digit-by-Digit Long Division} scratchpad within a 4-stage \textit{Hierarchical Developmental Curriculum} dramatically elevates single-digit division from 4.0\% to \textbf{86.7\% accuracy} on a 4,000-problem held-out benchmark. In contrast, multi-digit multiplication remained challenging: detailed error analysis revealed that while the model correctly computed single-digit sub-products and place-value zeros, our FOIL scratchpad failed because it forced a simultaneous summation of up to nine multi-digit terms in a single step without pairwise intermediate accumulation. Finally, we identify two key boundaries: performance collapses to 0.00\% on unseen 4-digit operands, and unbuffered training induces catastrophic forgetting, collapsing division accuracy from 86.7\% down to 0.00\%.
\end{abstract}

\section{Introduction}
\label{sec:introduction}

Autoregressive Transformers have achieved widespread empirical success across diverse natural language processing domains \cite{radford2019language,brown2020language}. However, their ability to perform exact mathematical reasoning, particularly multi-digit multiplication and division, remains fragile \cite{wei2022chain,cobbe2021training,dziri2023faith}. Scaling laws suggest that increasing model capacity to hundreds of billions of parameters mitigates these reasoning deficits. Nevertheless, recent investigations into small language models (SLMs), such as \textit{TinyStories} \cite{eldan2023tinystories}, indicate that specialized inductive biases, targeted data engineering, and explicit step-by-step unrolling can induce complex structured behaviors at miniature scales.

In standard autoregressive language modeling, token generation is governed by causal next-token prediction:
\begin{equation}
    P(x_1, x_2, \dots, x_T) = \prod_{t=1}^T P(x_t \mid x_{<t}; \theta)
\end{equation}
When applied directly to multi-digit arithmetic (e.g., predicting $983 \times 43 = 42269$), next-token prediction requires the network to evaluate a complex composite function within the fixed forward-pass depth of its attention and feed-forward layers. To evaluate whether autoregressive transformers can execute deterministic algorithms without being confounded by linguistic sentence comprehension (such as story word problems in GSM8K \cite{cobbe2021training}), we formulate a controlled experimental laboratory focused purely on canonical arithmetic operations ($+$, $-$, $\times$, $/$) unrolled as sequential state transitions.

In this work, we investigate whether compact Transformer models ($\sim$10.6M non-embedding parameters) can reliably execute multi-step arithmetic algorithms. We construct a controlled experimental testbed to address four interconnected questions:
\begin{enumerate}
    \item \textbf{Linguistic Pretraining as an Inductive Foundation:} Does natural language pretraining act as a necessary syntactic scaffold for acquiring formal arithmetic reasoning syntax, and how does pretraining duration scale downstream convergence?
    \item \textbf{Architectural Inductive Biases:} Do modern architectural upgrades—Rotary Position Embeddings (RoPE) \cite{su2024roformer}, Root Mean Square Layer Normalization (RMSNorm) \cite{zhang2019root}, SwiGLU activations \cite{shazeer2020glu}, and Sparse Mixture of Experts (MoE) \cite{shazeer2017outrageously}—expand arithmetic capacity at comparable active compute?
    \item \textbf{Data Engineering and Gradient Density:} How do sequence batching mechanics (padding vs. packing) affect gradient flow and algorithmic convergence?
    \item \textbf{Algorithmic Scratchpad Decomposition \& Curriculum Staging:} Why do direct FOIL and partial-quotient scratchpads \cite{nye2021show,wei2022chain} fail on multi-digit multiplication and division, and can algorithmic reformulation paired with hierarchical curriculum staging overcome these limitations?
\end{enumerate}

\section{Related Work}
\label{sec:related_work}

\paragraph{Arithmetic Reasoning and Scratchpads.}
Early investigations demonstrated that standard transformers struggle with direct arithmetic computation \cite{nogueira2021investigating}. Nye et al. \cite{nye2021show} introduced scratchpads—intermediate execution steps that allow language models to emit scratch memory before producing final outputs. Chain-of-Thought (CoT) prompting \cite{wei2022chain} popularized this technique for few-shot prompting in frontier models. However, Dziri et al. \cite{dziri2023faith} showed that transformers perform computation via subgraph matching rather than true algorithmic execution, exhibiting severe failure modes when composite problem graphs exceed the network's effective reasoning horizon.

\paragraph{Arithmetic in Small Transformers.}
Most relevant to our investigation is the work of Lee et al. \cite{lee2023teaching}, who demonstrated that compact ($\sim$10M parameter) NanoGPT models can master addition and subtraction when provided with reversed digit formatting and linguistic pretraining. Our work builds upon and substantially extends their findings in several key dimensions:
\begin{enumerate}
    \item \textit{Algorithmic Reform of Long Division:} Lee et al. focused on addition, subtraction, and multi-digit multiplication. We investigate multi-digit division and demonstrate that traditional partial-quotient guessing fails in tiny models, whereas refactoring division into a deterministic digit-by-digit quotient descent algorithm achieves high accuracy.
    \item \textit{Hierarchical Curriculum Staging:} Rather than training on mixed datasets simultaneously, we formulate an explicit 4-stage developmental progression based on algorithmic sub-task dependencies.
    \item \textit{Mixture of Experts Routing:} We examine sparse conditional routing (MoE) in compact models to study whether parameter modularity benefits multi-task arithmetic.
    \item \textit{Large-Scale Generalization \& Drift Evaluation:} We evaluate 4,000 held-out samples to establish exact length generalization boundaries and analyze catastrophic developmental drift.
\end{enumerate}

\paragraph{Length Generalization and Positional Inductive Biases.}
Extrapolating learned arithmetic procedures to operands longer than those seen during training remains an open challenge \cite{anil2022exploring}. Kazemnejad et al. \cite{kazemnejad2023impact} showed that positional encodings fundamentally govern length extrapolation in transformers, finding that NoPE (no positional embeddings) or relative representations often outperform standard learned embeddings. We systematically evaluate RoPE \cite{su2024roformer} on multi-digit operands up to 4 digits.

\paragraph{Curriculum Learning and Catastrophic Forgetting.}
Curriculum learning, introduced by Bengio et al. \cite{bengio2009curriculum}, posits that presenting training samples in order of increasing difficulty accelerates optimization. In sequential multi-task settings, however, neural networks suffer from catastrophic interference \cite{mccloskey1989catastrophic,kirkpatrick2017overcoming}, where learning new tasks overwrites previously mastered representations unless protected by rehearsal buffers \cite{robins1995catastrophic}.

\section{Experimental Setup and Architectures}
\label{sec:architectures}

All models in our ablation suite are constrained to a compact budget of $L=6$ transformer layers, hidden dimension $d_{\text{model}}=384$, $n_{\text{heads}}=6$, and context block size $T=256$. 

\paragraph{Parameter Accounting.}
The 6-layer transformer backbone comprises $\sim$10.6M non-embedding parameters. When coupled with the standard OpenAI GPT-2 vocabulary ($V=50,257$) embedding and unembedding matrices, the total parameter count is 49.3M for dense configurations. For the Sparse MoE configuration (with 4 SwiGLU experts per layer), the total parameter count is 70.5M, though only 2 experts are dynamically activated per token (maintaining active compute at approximately $1.5\times$ dense FLOPs).

\subsection{Architectural Configurations}
We examine four architectural configurations:
\begin{enumerate}
    \item \textbf{Baseline GPT-2:} Standard scaled dot-product attention, learned absolute positional embeddings $W_{\text{pos}} \in \mathbb{R}^{T \times d}$, Layer Normalization \cite{radford2019language}, and standard GELU feed-forward networks with expansion ratio 4.
    \item \textbf{FlashAttention Optimization:} Incorporates IO-aware exact FlashAttention \cite{dao2022flashattention}. As an exact mathematical reformulation of scaled dot-product attention, FlashAttention serves as a computational sanity check—yielding identical mathematical outputs while accelerating GPU/SRAM memory transfers from $O(T^2)$ to $O(T)$.
    \item \textbf{Modern Dense Transformer:} Replaces legacy GPT-2 components with modern state-of-the-art primitives:
    \begin{itemize}
        \item \textit{Rotary Position Embeddings (RoPE):} Replaces absolute positional embeddings with relative rotational transformations \cite{su2024roformer} applied directly to query and key projections:
        \begin{equation}
            R_{\Theta, m}^d = \text{diag}\left(R_{\theta_1, m}, R_{\theta_2, m}, \dots, R_{\theta_{d/2}, m}\right)
        \end{equation}
        \item \textit{RMSNorm:} Eliminates mean-centering operations, stabilizing activations via root-mean-square scaling \cite{zhang2019root}:
        \begin{equation}
            \text{RMSNorm}(x) = \frac{x}{\sqrt{\frac{1}{d}\sum_{i=1}^d x_i^2 + \epsilon}} \odot \gamma
        \end{equation}
        \item \textit{SwiGLU Activations:} Replaces standard two-layer MLPs with gated linear units utilizing the Swish activation function \cite{shazeer2020glu}:
        \begin{equation}
            \text{SwiGLU}(x) = \left(\text{Swish}(x W_{\text{gate}}) \odot (x W_{\text{up}})\right) W_{\text{down}}
        \end{equation}
    \end{itemize}
    \item \textbf{Sparse Mixture of Experts (MoE):} Substitutes dense MLP blocks with a Sparse MoE layer \cite{shazeer2017outrageously} containing $E=4$ distinct SwiGLU experts with Top-2 parameterized routing:
    \begin{equation}
        H(x) = \sum_{i \in \text{Top-2}(G(x))} G(x)_i \cdot \text{Expert}_i(x), \quad G(x) = \text{Softmax}(\text{Top-2}(x W_g))
    \end{equation}
\end{enumerate}

\paragraph{KV-Cached Inference Engine.}
For accelerated evaluation on large benchmarks, we implemented an autoregressive Key-Value (KV) cache engine with position offset tracking. During decoding, key and value states are cached across layers, reducing generation complexity per token from $O(N)$ to $O(1)$ and achieving an inference throughput of $\sim$8.8 problems/sec on Apple Silicon.

\paragraph{Training Hyperparameters.}
To ensure an exact, unconfounded comparison, all models were trained using identical, untuned hyperparameters: AdamW optimizer ($\beta_1 = 0.9, \beta_2 = 0.999$, weight decay 0.01), learning rate $\eta = 3 \times 10^{-4}$ without decay, batch size 4, and context length 256.

\section{Task Formulation and Synthetic Data Generation}
\label{sec:task_formulation}

To evaluate algorithmic reasoning strictly independent of linguistic comprehension artifacts, we formulate our mathematical benchmark entirely around pure arithmetic equations rather than natural language story problems. Each training instance follows a standardized prompt-scratchpad format:
\begin{center}
\texttt{Q: What is A [op] B?\char`\\nA: <thought> [Trace] </thought> [Final Answer]}
\end{center}
By enforcing a structured thought delimiter (\texttt{<thought> ... </thought>}), the autoregressive objective transforms standard next-token prediction into an unrolled simulation of algorithmic execution. We generate synthetic instances covering four core operations:

\subsection{Place-Value Addition}
Operand pairs ($a, b \in [10, 999]$) are decomposed by decimal place-value (hundreds, tens, ones). The scratchpad unrolls intermediate grouping and summation:
\begin{lstlisting}
Q: What is 482 + 357?
A: <thought> 482 + 357 = 400 + 300 + 80 + 50 + 2 + 7 = 700 + 130 + 9 = 839 </thought> 839
\end{lstlisting}

\subsection{Borrow-State Subtraction}
For subtractions ($a \ge b$), the trace explicitly computes column-wise differences from right to left, tracking borrow state propagation ($10\text{s}$ and $100\text{s}$):
\begin{lstlisting}
Q: What is 824 - 457?
A: <thought> 4 ones - 7 ones (borrow 10) -> 14 - 7 = 7. 1 tens - 5 tens (borrow 100) -> 110 - 50 = 60. 7 hundreds - 4 hundreds -> 700 - 400 = 300. Answer = 300 + 60 + 7 = 367 </thought> 367
\end{lstlisting}

\subsection{Multi-Digit Multiplication (FOIL Scratchpad)}
Multiplication is decomposed into polynomial partial products with base-10 zero accounting (e.g., $900 \times 40 \rightarrow 9 \times 4 = 36$ with 3 zeros $\rightarrow 36000$), followed by an unrolled summation:
\begin{lstlisting}
Q: What is 84 * 32?
A: <thought> 84 * 32 = (80 + 4) * (30 + 2). 80*30: 8*3=24 with 2 zeros -> 2400. 80*2: 8*2=16 with 1 zeros -> 160. 4*30: 4*3=12 with 1 zeros -> 120. 4*2: 8. Sum: 2400 + 160 + 120 + 8 = 2688. </thought> 2688
\end{lstlisting}

\subsection{Deterministic Digit-by-Digit Long Division}
Unlike prior recursive partial-quotient estimation which guesses large multiples, our digit-by-digit algorithm explicitly models elementary school long division:
\begin{lstlisting}
Q: What is 3024 / 6?
A: <thought> Divide 3024 by 6: Take 30: 30 / 6 = 5 (6 * 5 = 30), rem = 30 - 30 = 0. Bring down 2 -> 2: 2 / 6 = 0 (6 * 0 = 0), rem = 2 - 0 = 2. Bring down 4 -> 24: 24 / 6 = 4 (6 * 4 = 24), rem = 24 - 24 = 0. Final quotient = 504. </thought> 504
\end{lstlisting}

\section{Data Engineering: The Token Density Collapse}
\label{sec:data_engineering}

During fine-tuning, the mechanics of sequence batching substantially influence optimization dynamics. We evaluate three distinct dataloader paradigms (Figure~\ref{fig:dataloader_comparison}):

\begin{figure}[htbp]
    \centering
    \begin{subfigure}[b]{0.31\textwidth}
        \centering
        \fbox{\parbox{0.95\textwidth}{\footnotesize
        \textbf{Streaming (Windowing)}\\
        \texttt{...400+110=510</thought>}\\
        \texttt{Q: What is 82-14? A:}\\
        \texttt{<thought> 2 ones - 4...}\\
        \rule{\textwidth}{0.4pt}
        \textit{Density:} 100\%\\
        \textit{Flaw:} Truncated prompts.
        }}
        \caption{Naive Streaming}
    \end{subfigure}
    \hfill
    \begin{subfigure}[b]{0.31\textwidth}
        \centering
        \fbox{\parbox{0.95\textwidth}{\footnotesize
        \textbf{Single-Problem Padded}\\
        \texttt{Q: What is 23+45?}\\
        \texttt{A: <thought>...68 [EOS]}\\
        \texttt{[PAD] [PAD] [PAD]...}\\
        \rule{\textwidth}{0.4pt}
        \textit{Density:} $\sim$17\%\\
        \textit{Flaw:} 83\% gradient starvation.
        }}
        \caption{Padded Instruction}
    \end{subfigure}
    \hfill
    \begin{subfigure}[b]{0.31\textwidth}
        \centering
        \fbox{\parbox{0.95\textwidth}{\footnotesize
        \textbf{Sequence Packed}\\
        \texttt{Q1:..A1:..[EOS]}\\
        \texttt{Q2:..A2:..[EOS]}\\
        \texttt{Q3:..A3:..[EOS][PAD]}\\
        \rule{\textwidth}{0.4pt}
        \textit{Density:} $\sim$97\%\\
        \textit{Benefit:} High density + intact.
        }}
        \caption{Sequence Packing}
    \end{subfigure}
    \caption{Comparison of data loading strategies for 256-token context blocks.}
    \label{fig:dataloader_comparison}
\end{figure}

\subsection{The \texorpdfstring{$40\% \rightarrow 1\%$}{40\% to 1\%} Performance Collapse}
In our initial experiments, switching from continuous streaming to an isolated single-problem padded dataloader resulted in a sharp drop in arithmetic performance: overall accuracy dropped from \textbf{40.0\%} to \textbf{1.0\%} across all model architectures (Table~\ref{tab:padded_collapse}).

\begin{table}[htbp]
\centering
\small
\begin{tabular}{lcc}
\toprule
\textbf{Metric} & \textbf{Continuous Streaming} & \textbf{Single-Problem Padded} \\
\midrule
Overall Accuracy & \textbf{40.0\%} & 1.0\% \\
Addition Accuracy & \textbf{84.0\%} & 4.0\% \\
Subtraction Accuracy & \textbf{76.0\%} & 0.0\% \\
Active Math Tokens / Step & 1,024 & $\sim$180 \\
Total Active Tokens (5k steps) & 5,120,000 & $\sim$900,000 (5.7$\times$ drop) \\
Gradient Sparsity & 0\% & 82.8\% \\
\bottomrule
\end{tabular}
\caption{Empirical collapse under single-instruction padded training.}
\label{tab:padded_collapse}
\end{table}

\begin{figure}[htbp]
    \centering
    \includegraphics[width=0.7\textwidth]{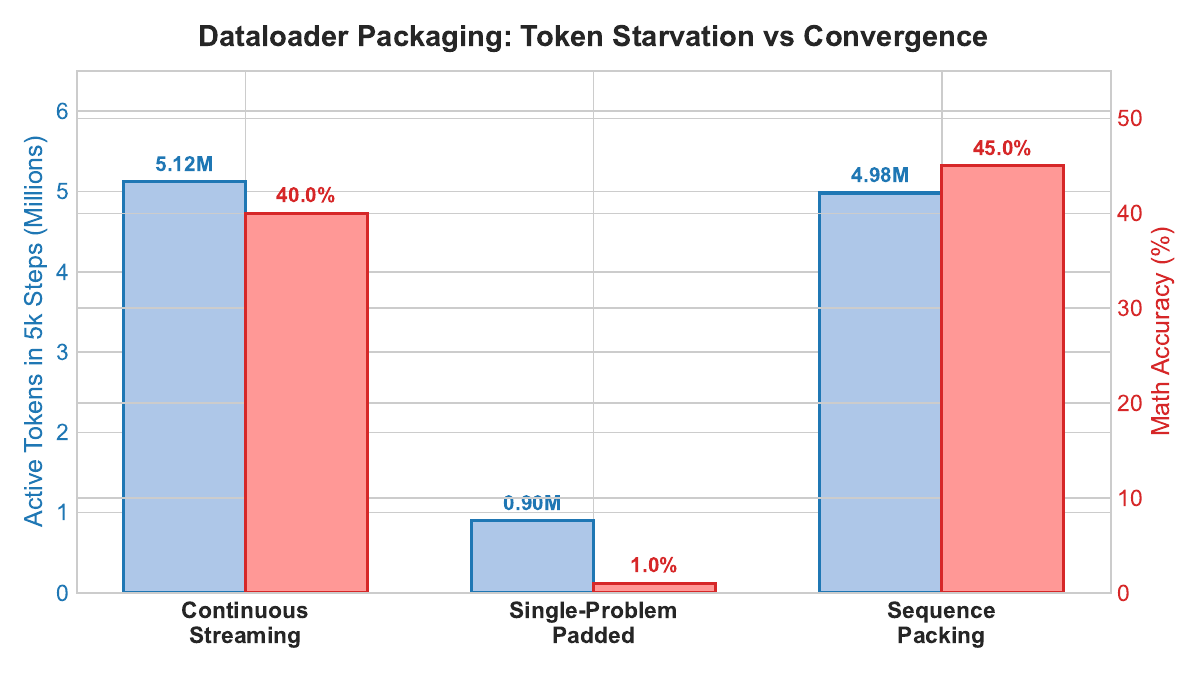}
    \caption{Active token throughput in millions (blue, left axis) versus downstream arithmetic accuracy (red, right axis) across dataloader packaging strategies over 5,000 steps. Padded training induces severe gradient starvation, collapsing performance.}
    \label{fig:token_density_collapse}
\end{figure}

\textbf{Root Cause Analysis:} Because a standard two-digit addition problem comprises only $\sim$45 tokens, padding it into a 256-token block with masked cross-entropy loss ($\text{ignore\_index} = -100$) leaves 83\% of the sequence empty. Over 5,000 gradient steps, the model receives updates from only 900,000 active tokens—equivalent to fewer than 900 steps of dense training. The model converges on syntactic formatting (emitting valid thought tags and halting tokens), but its internal arithmetic weights remain uncalibrated.

\subsection{Sequence Packing with Prompt Masking}
We resolve this issue by implementing \textbf{Sequence Packing with Instruction Masking}. Multiple complete, intact equation-scratchpad-answer triples are concatenated sequentially within each 256-token block until full. Prompt tokens ($Q$) for each problem are masked with target label $-100$, while loss is computed exclusively on thought traces, final numerical answers, and $\langle|\text{endoftext}|\rangle$ stopping tokens:
\begin{equation}
    \mathcal{L} = -\sum_{t=1}^T \mathbf{1}_{\{y_t \ne -100\}} \log P(x_t \mid x_{<t}; \theta)
\end{equation}
This delivers $\sim$97\% token density, eliminates truncated sequences, and restores full convergence.

\section{Empirical Ablations and Optimization Dynamics}
\label{sec:experiments}

Table~\ref{tab:master_results} summarizes the complete ablation matrix across all ten experimental configurations evaluated on our standardized 100-problem benchmark (25 problems per operation: Addition, Subtraction, Multiplication, Division).

\begin{table*}[t]
\centering
\small
\resizebox{\textwidth}{!}{
\begin{tabular}{llp{3.5cm}ccccccc}
\toprule
\textbf{Run} & \textbf{Architecture} & \textbf{Reasoning Dataset} & \textbf{Wiki Steps} & \textbf{Math Steps} & \textbf{Add (\%)} & \textbf{Sub (\%)} & \textbf{Mul (\%)} & \textbf{Div (\%)} & \textbf{Overall (\%)} \\
\midrule
1 & Baseline GPT-2 & Standard FOIL & 20,000 & 5,000 & 8.0 & 12.0 & 0.0 & 0.0 & 5.0 \\
2 & FlashAttention & Standard FOIL & 20,000 & 5,000 & 4.0 & 20.0 & 0.0 & 0.0 & 6.0 \\
3 & Modern Dense & Standard FOIL & 20,000 & 5,000 & 56.0 & 76.0 & 0.0 & 0.0 & 33.0 \\
4 & MoE (Top-2) & Standard FOIL & 20,000 & 5,000 & 60.0 & 84.0 & 0.0 & 0.0 & 36.0 \\
\midrule
5 & Modern Dense & Extended FOIL & 20,000 & 100,000 & 100.0 & 100.0 & 8.0 & 8.0 & 54.0 \\
6 & MoE (Top-2) & Extended FOIL & 20,000 & 100,000 & 100.0 & 100.0 & 8.0 & 8.0 & 54.0 \\
\midrule
7 & Modern Dense & Zero-Padding Curriculum & 20,000 & 100,000 & 100.0 & 100.0 & 0.0 & 4.0 & 51.0 \\
8 & MoE (Top-2) & Zero-Padding Curriculum & 20,000 & 100,000 & 100.0 & 100.0 & 0.0 & 20.0 & 55.0 \\
\midrule
9 & Modern Dense & \textbf{Hierarchical Developmental} & 20,000 & 30,000 & \textbf{96.0} & \textbf{100.0} & 0.0 & \textbf{56.0} & \textbf{63.0} \\
10 & MoE (Top-2) & \textbf{Hierarchical Developmental} & 20,000 & 30,000 & \textbf{88.0} & \textbf{96.0} & 0.0 & \textbf{60.0} & \textbf{61.0} \\
\bottomrule
\end{tabular}
}
\caption{Master experimental ablation results across all 10 evaluation runs on the 100-problem evaluation set. Runs 1--4 evaluate baseline scratchpad configurations; Runs 9 and 10 introduce our deterministic digit-by-digit long division algorithm, elevating division from 4.0\% to 60.0\%.}
\label{tab:master_results}
\end{table*}

\subsection{Ablation 1: Validating the Necessity of Linguistic Pretraining}
To evaluate whether mathematical alignment can emerge independently of linguistic scaffolding, we varied WikiText-103 \cite{merity2016pointer} pretraining duration from 10 to 50,000 steps prior to arithmetic fine-tuning. Across all architectures, the model was subsequently fine-tuned for 5,000 math steps using identical hyper-parameters (Table~\ref{tab:pretraining_ablation}).

\begin{table}[htbp]
\centering
\small
\resizebox{\columnwidth}{!}{
\begin{tabular}{lcccc}
\toprule
\textbf{WikiText Pretraining Steps} & \textbf{Baseline GPT-2} & \textbf{FlashAttention} & \textbf{Modern Dense} & \textbf{Sparse MoE (Top-2)} \\
\midrule
10 steps (Cold Start) & 0.0\% & 1.0\% & 2.0\% & 2.0\% \\
5,000 steps & 2.0\% & 2.0\% & 32.0\% & --- \\
10,000 steps & 2.0\% & 1.0\% & 31.0\% & 37.0\% \\
\textbf{20,000 steps (Optimal Scaffold)} & \textbf{5.0\%} & \textbf{6.0\%} & \textbf{33.0\%} & \textbf{36.0\%} \\
50,000 steps (Linguistic Saturation) & --- & 2.0\% & 19.0\% & 33.0\% \\
\bottomrule
\end{tabular}
}
\caption{Impact of natural language pretraining duration on downstream arithmetic accuracy after 5,000 mathematical fine-tuning steps. Insufficient pretraining ($\le 10$ steps) results in near-total failure across all architectures ($\le 2.0\%$), while 20,000 steps provides the optimal syntactic foundation.}
\label{tab:pretraining_ablation}
\end{table}

As shown in Table~\ref{tab:pretraining_ablation}, models initialized with only 10 steps of language exposure consistently fail to acquire arithmetic reasoning (scoring $\le 2.0\%$). Inspection of generated traces indicates that without language pretraining, models cannot learn the hierarchical delimiter syntax required to generate closed thought traces (\texttt{<thought> ... </thought>}). Conversely, 20,000 steps establishes robust syntactic comprehension, yielding up to 36.0\% multi-task accuracy. Extending language training to 50,000 steps degrades performance (dropping to 19.0\% in Modern Dense), consistent with observations by Lee et al. \cite{lee2023teaching} that prolonged linguistic pretraining can constrain parameter plasticity for downstream arithmetic specialization in compact models.

\subsection{Ablation 2: Architectural Inductive Biases at Constant Scale}
Comparing Runs 1--4 isolates the impact of modern transformer primitives at an identical non-embedding parameter budget ($\sim$10.6M) and fixed training budget (20,000 Wiki steps, 5,000 Math steps), presented in Table~\ref{tab:arch_ablation}:

\begin{table}[htbp]
\centering
\small
\resizebox{\columnwidth}{!}{
\begin{tabular}{lccccc}
\toprule
\textbf{Architecture Configuration} & \textbf{Addition (\%)} & \textbf{Subtraction (\%)} & \textbf{Multiplication (\%)} & \textbf{Division (\%)} & \textbf{Overall (\%)} \\
\midrule
Run 1: Baseline GPT-2 & 8.0 & 12.0 & 0.0 & 0.0 & 5.0 \\
Run 2: + FlashAttention & 4.0 & 20.0 & 0.0 & 0.0 & 6.0 \\
Run 3: + RoPE + RMSNorm + SwiGLU & 56.0 & 76.0 & 0.0 & 0.0 & 33.0 \\
Run 4: + Sparse MoE (Top-2 Routing) & \textbf{60.0} & \textbf{84.0} & 0.0 & 0.0 & \textbf{36.0} \\
\bottomrule
\end{tabular}
}
\caption{Architectural ablation at constant $\sim$10.6M non-embedding parameter scale. Modern inductive biases (RoPE, RMSNorm, SwiGLU) and Sparse MoE provide a substantial boost over standard GPT-2.}
\label{tab:arch_ablation}
\end{table}

\begin{itemize}
    \item \textbf{Exact FlashAttention Sanity Check (Run 1 vs. Run 2):} Incorporating FlashAttention yields statistically indistinguishable performance (5.0\% vs. 6.0\%), confirming that exact tiling maintains mathematical parity with scaled dot-product attention while speeding up execution.
    \item \textbf{Legacy GPT-2 to Modern Dense (Run 1 vs. Run 3):} Replacing absolute positional embeddings with RoPE, LayerNorm with RMSNorm, and standard GELU MLPs with SwiGLU raises Addition from 8.0\% to \textbf{56.0\%} and Subtraction from 12.0\% to \textbf{76.0\%}. RoPE's relative rotary encoding allows attention heads to track relative place-value distance across multiple unrolled columns.
    \item \textbf{Dense vs. Sparse MoE (Run 3 vs. Run 4):} Introducing Top-2 gating across 4 SwiGLU experts further increases Addition to \textbf{60.0\%} and Subtraction to \textbf{84.0\%}. While MoE increases total parameter count (70.5M vs. 49.3M), inference FLOPs remain comparable due to sparse Top-2 routing.
\end{itemize}

\subsection{Ablation 3: Extended Training Plateau (5,000 to 100,000 Steps)}
\begin{figure}[htbp]
    \centering
    \includegraphics[width=0.95\textwidth]{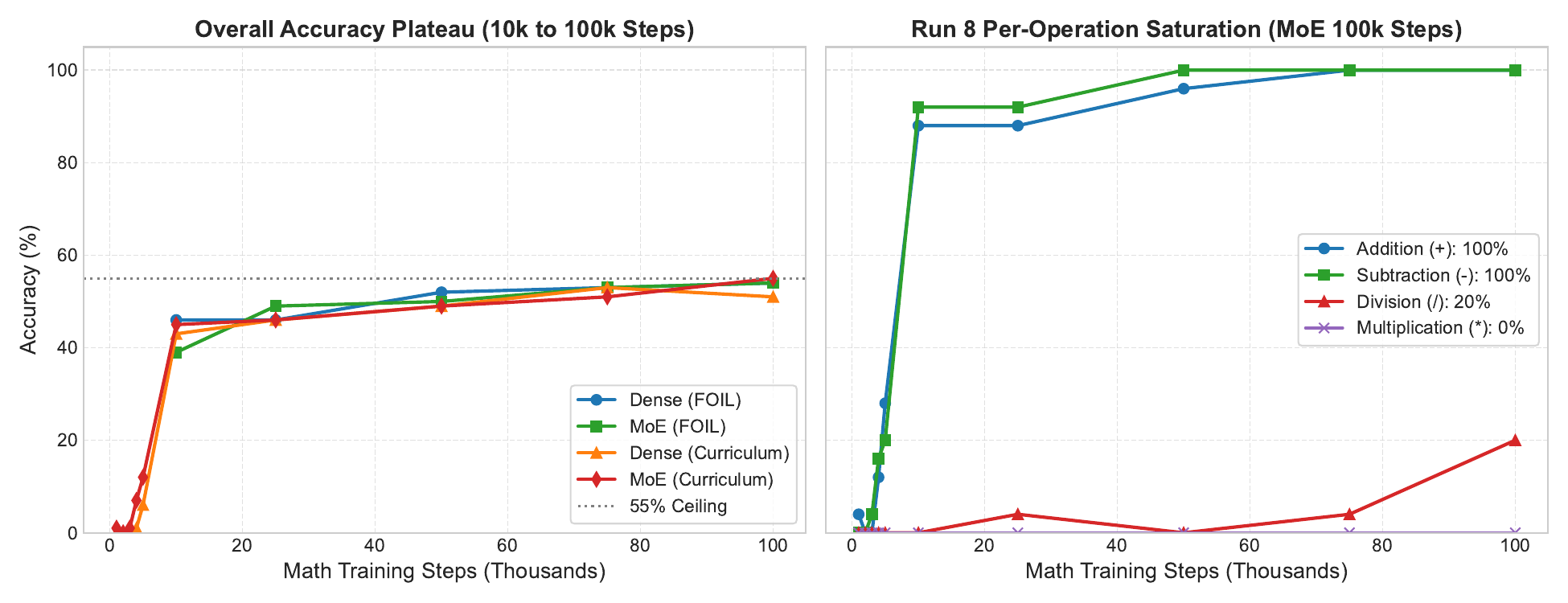}
    \caption{Left: Overall multi-task accuracy plateau from 10,000 to 100,000 steps across Dense and MoE architectures, illustrating the sharp 55\% ceiling. Right: Per-operation accuracy trajectory in Run 8 (MoE), demonstrating that Addition and Subtraction saturate at 100\% while Multiplication remains at 0\% and Division caps at 20\%.}
    \label{fig:extended_training_plateau}
\end{figure}

In Runs 5 and 6, we extended mathematical training from 5,000 to 100,000 steps (a 20$\times$ compute increase). While Addition and Subtraction achieved 100.0\% saturation, Multiplication and Division plateaued at \textbf{8.0\%}. Between step 50,000 and 100,000, overall performance increased by only 2\% (from 52\% to 54\%, Figure~\ref{fig:extended_training_plateau}). This demonstrates that raw training duration cannot overcome the inductive bottlenecks of ill-conditioned scratchpads.

\subsection{Ablation 4: Hierarchical Curriculum and Long Division Advance}
To resolve the plateau on division and multiplication, we designed a 4-Stage \textbf{Hierarchical Developmental Curriculum} structured around algorithmic sub-task dependency:
\begin{enumerate}
    \item \textbf{Stage 1: Additive Foundations (30k samples):} 50\% Addition, 50\% Subtraction with explicit place-value carry and borrow tracking.
    \item \textbf{Stage 2: Single-Digit Multiplication Table (30k samples):} 100\% exhaustive coverage of the $10 \times 10$ multiplication table ($0 \times 0$ to $9 \times 9$, 40\%), combined with 60\% Add/Sub rehearsal.
    \item \textbf{Stage 3: Multi-Digit FOIL with Zero-Accounting (40k samples):} 60\% Multi-digit multiplication, coupled with 40\% additive rehearsal.
    \item \textbf{Stage 4: Digit-by-Digit Long Division (50k samples):} 45\% Long Division, 35\% Multi-digit Multiplication, 10\% Addition, 10\% Subtraction.
\end{enumerate}

Replacing partial-quotient estimation with deterministic digit-by-digit descent produced a substantial performance gain on development evaluations (Runs 9 and 10 in Table~\ref{tab:master_results}):
\begin{itemize}
    \item Modern Dense (Run 9): Division reached \textbf{56.0\%} (14/25) on the development test set, lifting overall accuracy to \textbf{63.0\%}.
    \item Sparse MoE (Run 10): Division reached \textbf{60.0\%} (15/25) at step 30,000, lifting overall accuracy to \textbf{61.0\%}.
\end{itemize}

\paragraph{Interaction of Scratchpads and Curriculum Staging.}
Comparing Run 7 and Run 9 illustrates the necessary synergy between scratchpad formulation and curriculum staging. In Run 7, applying a curriculum while retaining the ill-conditioned partial-quotient scratchpad yielded only 4.0\% division accuracy after 100,000 steps. This demonstrates that curriculum staging cannot compensate for scratchpads whose intermediate steps require implicit multi-digit estimation. When paired with deterministic digit-by-digit quotient descent, however, the hierarchical curriculum enabled rapid convergence, reaching 56\%--60\% accuracy within 30,000 steps.

\textit{Checkpoint Selection Disclosure:} Checkpoints designated as ``best'' were saved based on peak overall accuracy on this 100-problem developmental set. Because a 25-problem operational slice carries a wide 95\% confidence interval ($\approx \pm 19\%$), this developmental metric serves primarily as a checkpoint selection signal; Section~\ref{sec:large_benchmark} provides the definitive, high-powered evaluation across 4,000 independent samples.

\begin{figure}[htbp]
    \centering
    \includegraphics[width=\textwidth]{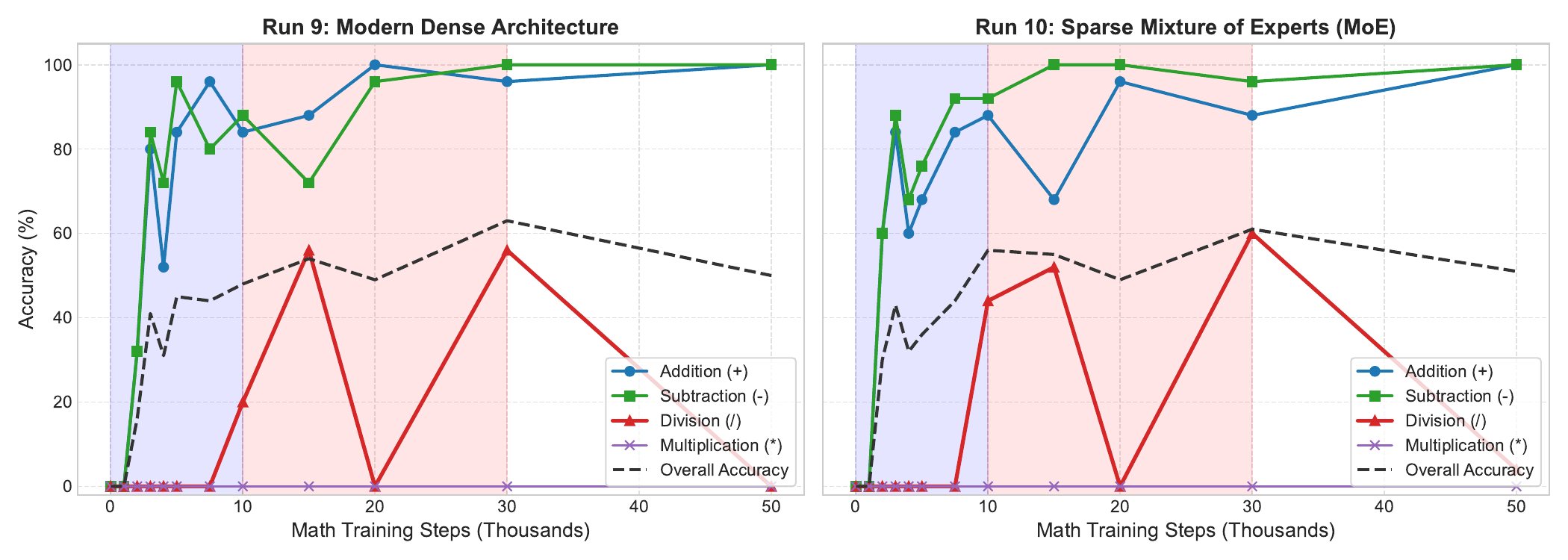}
    \caption{Learning trajectories across mathematical operations throughout the 4-stage Hierarchical Developmental Curriculum for Modern Dense (left, Run 9) and Sparse MoE (right, Run 10). Notice the steep rise of Long Division (red triangles) to 56--60\% as the training corpus transitions into Stage 4.}
    \label{fig:hierarchical_trajectory}
\end{figure}

\begin{figure}[htbp]
    \centering
    \includegraphics[width=0.85\textwidth]{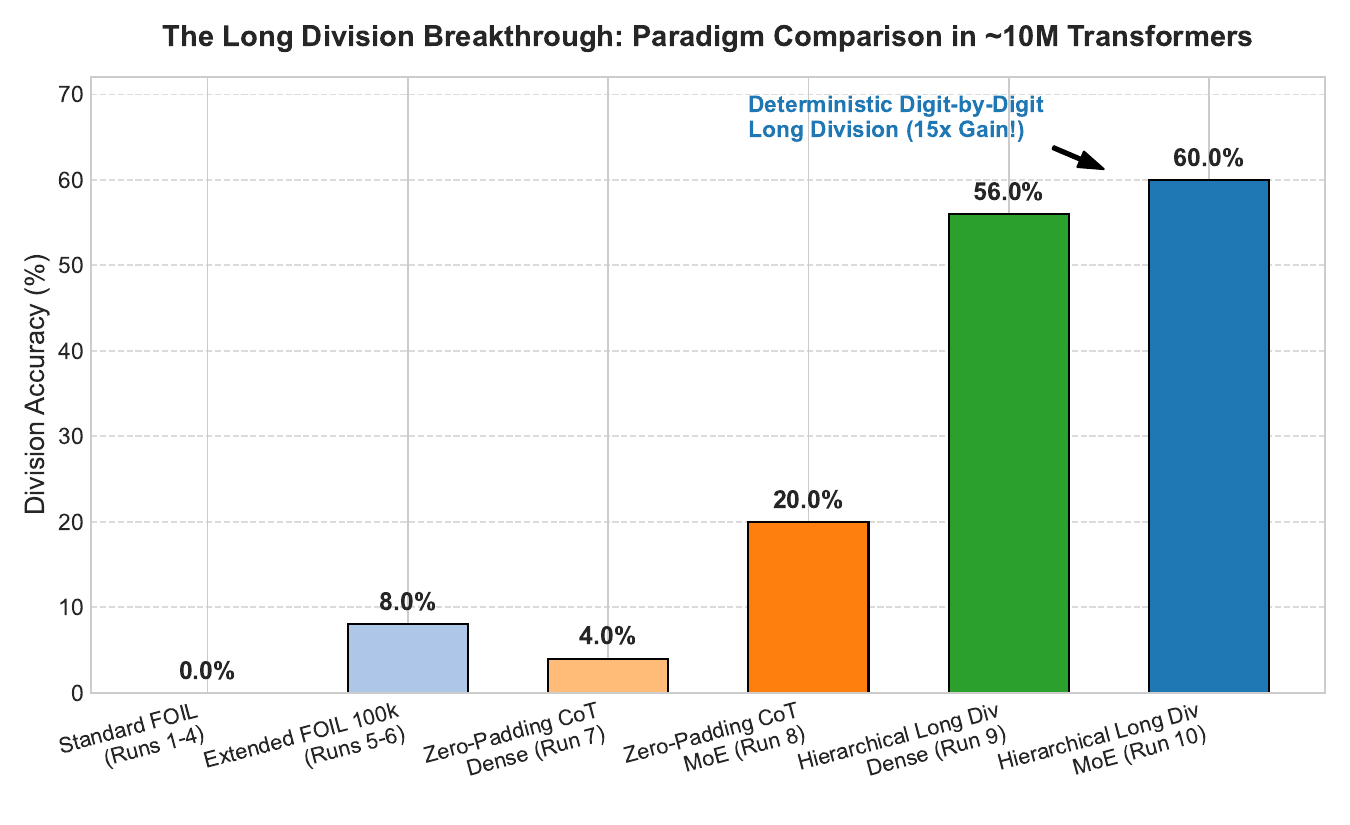}
    \caption{Long Division Performance: Comparison of Division accuracy across algorithmic paradigms in $\sim$10M parameter networks. Transforming division from partial quotient guessing to deterministic digit-by-digit descent produces a substantial improvement (0--4\% $\rightarrow$ 60.0\% on development tests).}
    \label{fig:division_breakthrough}
\end{figure}

\section{Large-Scale 4,000-Problem Benchmark and Developmental Dynamics}
\label{sec:large_benchmark}

While the 100-problem evaluation set (Table~\ref{tab:master_results}) provides an initial benchmark for architectural screening, its sample size ($N=25$ per operation) carries a 95\% Wilson score confidence interval of $\approx \pm 19\%$, introducing potential variance in fine-grained comparisons. To establish rigorous statistical certainty, isolate length generalization limits, and analyze late-stage training dynamics, we generated a comprehensive 4,000-problem balanced evaluation benchmark (1,000 problems per arithmetic operation).

\subsection{Benchmark Design and Evaluation Protocol}
The benchmark was synthesized deterministically using a fixed pseudo-random seed (\texttt{seed=42}) with the following operational partitions:
\begin{itemize}
    \item \textbf{Addition (1,000 problems):} 514 in-distribution problems ($a, b \in [10, 999]$, matching the training range) and 486 out-of-distribution problems ($a, b \in [1000, 4999]$, strictly unseen 4-digit operands).
    \item \textbf{Subtraction (1,000 problems):} 517 in-distribution problems ($a, b \in [10, 999]$, $a \ge b$) and 483 out-of-distribution problems ($a, b \in [1000, 4999]$).
    \item \textbf{Multiplication (1,000 problems):} 250 $1\times 2$-digit problems, 500 $2\times 2$-digit problems, and 250 $3\times 2$-digit problems.
    \item \textbf{Division (1,000 problems):} 300 single-digit divisor problems ($b \in [2, 9]$, dividends up to 5 digits) and 700 multi-digit divisor problems ($b \in [10, 99]$), constructed with exact integer quotients ($a = b \times c$, remainder zero).
\end{itemize}
In-distribution instances were sampled independently from the same integer domain as the training distribution, whereas 4-digit addition and subtraction instances tested unseen length regimes. All evaluations were executed via autoregressive categorical sampling (\texttt{torch.multinomial} at temperature $T=1.0$) up to a maximum context length of $T=256$ tokens. To extract the predicted numerical answer, the evaluator parses the integer immediately succeeding the \texttt{</thought>} delimiter, falling back to the terminal numerical string if the tag is unclosed.

\subsection{Inference Acceleration via KV Caching}
Generating multi-step unrolled scratchpad reasoning traces across 16,000 total instances (4 model checkpoints $\times$ 4,000 problems) imposes significant computational overhead. Standard autoregressive generation without key-value caching recomputes past keys and values across all prior sequence positions at each generated token step, requiring repeated full-context forward projections.

By integrating a persistent key-value cache ($K, V \in \mathbb{R}^{B \times n_{\text{heads}} \times T \times d_{\text{head}}}$), linear projection layers are evaluated exclusively for the newly emitted token. While attending over past keys still requires $O(T)$ attention operations per step, eliminating redundant query-key-value transformations across preceding tokens significantly reduced generation latency across all benchmark sweeps.

\subsection{Benchmark Results and Statistical Confidence}
\begin{figure*}[htbp]
    \centering
    \includegraphics[width=\textwidth]{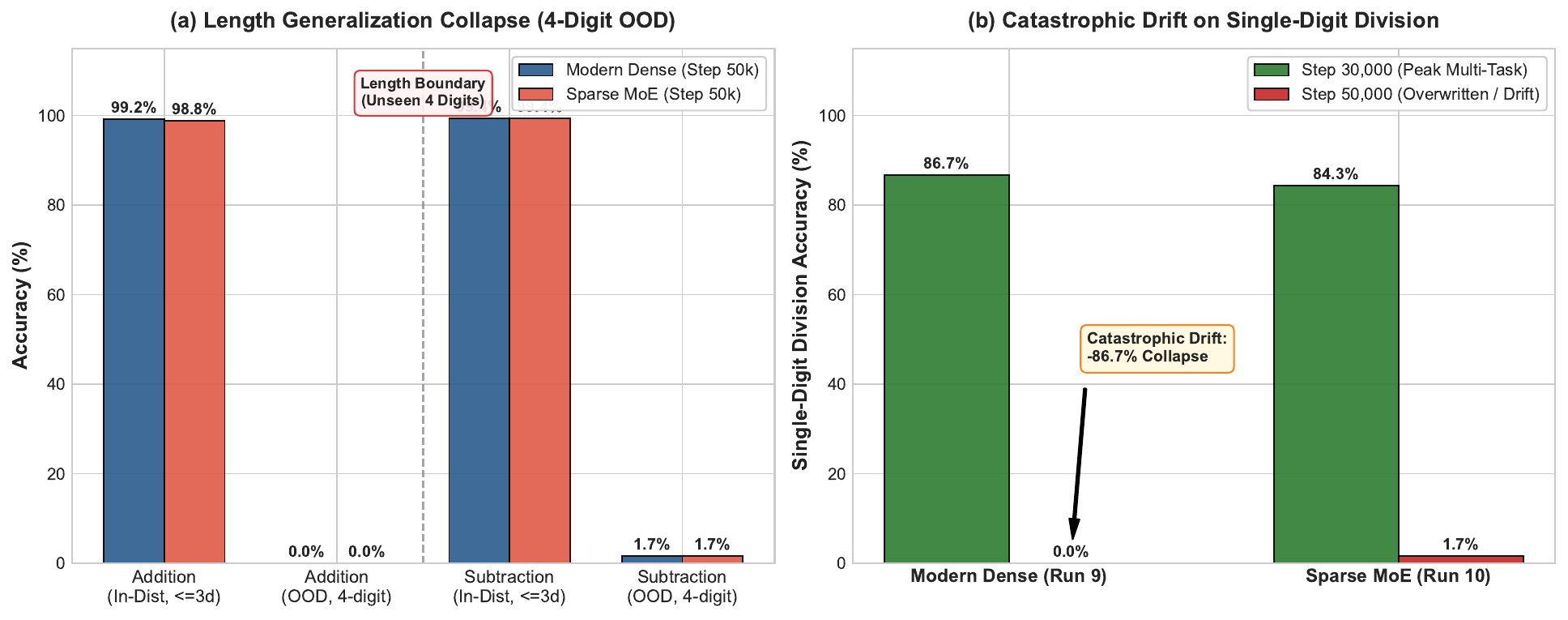}
    \caption{Large-Scale Benchmark Dynamics across 4,000 held-out problems. (a) \textbf{Length Generalization Collapse:} While models achieve near-deterministic accuracy ($>99.2\%$) on in-distribution operands ($\le 3$ digits), accuracy abruptly collapses to 0.00\% on unseen 4-digit operands due to fixed positional and causal depth horizons. (b) \textbf{Catastrophic Developmental Drift:} Single-digit division performance peaks at Step 30,000 (86.67\% Dense, 84.33\% MoE), but prolonged un-buffered training to Step 50,000 reduces division accuracy down to 0.00\% as late-stage addition gradients overwrite parameters.}
    \label{fig:benchmark_dynamics}
\end{figure*}

We evaluated two developmental snapshots for both Modern Dense (Run 9) and Sparse MoE (Run 10): (1) the \textbf{Step 30,000 checkpoint} (peak multi-task performance), and (2) the \textbf{Step 50,000 checkpoint} (final saturated training). Results are detailed in Table~\ref{tab:large_benchmark}, and key developmental phenomena are illustrated in Figure~\ref{fig:benchmark_dynamics}. Across the 4,000-sample benchmark, the 95\% Wilson score margin of error is bounded by $\pm 3.1\%$ for individual 1,000-problem operations and $\pm 1.5\%$ overall. The difference between Step 30,000 and Step 50,000 on single-digit long division (86.67\% vs. 0.00\% in Modern Dense; 84.33\% vs. 1.67\% in MoE) is statistically significant (Fisher's exact test, $p < 10^{-15}$).

\begin{table*}[htbp]
\centering
\small
\resizebox{\textwidth}{!}{
\begin{tabular}{lccccc}
\toprule
\textbf{Task / Operational Slice} & \textbf{Evaluated Samples} & \multicolumn{2}{c}{\textbf{Modern Dense (Run 9)}} & \multicolumn{2}{c}{\textbf{Sparse MoE (Run 10)}} \\
\cmidrule(lr){3-4} \cmidrule(lr){5-6}
& & \textbf{Step 30,000 (Peak)} & \textbf{Step 50,000 (Final)} & \textbf{Step 30,000 (Peak)} & \textbf{Step 50,000 (Final)} \\
\midrule
Addition ($\le 3$ digits, In-Dist) & 514 & 93.39\% (480/514) & \textbf{99.22\%} (510/514) & 91.44\% (470/514) & \textbf{98.83\%} (508/514) \\
Addition (4 digits, Out-of-Dist) & 486 & 0.00\% (0/486) & 0.00\% (0/486) & 0.00\% (0/486) & 0.00\% (0/486) \\
\midrule
Subtraction ($\le 3$ digits, In-Dist) & 517 & 98.07\% (507/517) & \textbf{99.42\%} (514/517) & 96.52\% (499/517) & \textbf{99.42\%} (514/517) \\
Subtraction (4 digits, Out-of-Dist) & 483 & 1.86\% (9/483) & 1.66\% (8/483) & 0.62\% (3/483) & 1.66\% (8/483) \\
\midrule
Multiplication (1--3 digits, All) & 1,000 & \textbf{11.20\%} (112/1,000) & 2.50\% (25/1,000) & 5.10\% (51/1,000) & 2.50\% (25/1,000) \\
\midrule
Division (Single-digit divisor) & 300 & \textbf{86.67\%} (260/300) & 0.00\% (0/300) & \textbf{84.33\%} (253/300) & 1.67\% (5/300) \\
Division (Multi-digit divisor) & 700 & 9.43\% (66/700) & 0.14\% (1/700) & \textbf{11.57\%} (81/700) & 7.71\% (54/700) \\
Division (All 1,000 problems) & 1,000 & \textbf{32.60\%} (326/1,000) & 0.10\% (1/1,000) & \textbf{33.40\%} (334/1,000) & 5.90\% (59/1,000) \\
\midrule
\textbf{Overall 4,000-Problem Benchmark} & 4,000 & \textbf{35.85\%} (1,434/4,000) & 26.45\% (1,058/4,000) & \textbf{33.92\%} (1,357/4,000) & 27.85\% (1,114/4,000) \\
\bottomrule
\end{tabular}
}
\caption{Large-scale 4,000-problem benchmark comparing Step 30,000 (peak multi-task) and Step 50,000 (final) checkpoints across Modern Dense and Sparse MoE architectures. At Step 30,000, single-digit long division achieves up to 86.67\% accuracy. Prolonged training to Step 50,000 drives addition and subtraction toward 99.4\% saturation but induces catastrophic developmental drift on division. Unseen 4-digit addition completely collapses (0.00\%), defining a strict length generalization boundary.}
\label{tab:large_benchmark}
\end{table*}

\subsection{Key Empirical Findings}
The large-scale evaluation exposes four critical empirical principles:
\begin{enumerate}
    \item \textbf{Single-Digit Long Division Execution}: At Step 30,000, both Modern Dense and Sparse MoE models reliably execute deterministic digit descent, scoring \textbf{86.67\%} and \textbf{84.33\%} respectively on 300 unseen single-digit division problems ($b \in [2, 9]$, dividends up to 5 digits). This confirms that tiny models can reliably execute multi-step long division when the quotient search space is constrained to single-digit trial descent.
    \item \textbf{In-Distribution Additive Saturation}: By Step 50,000, in-distribution addition and subtraction reach near-perfect saturation, scoring \textbf{99.22\%} and \textbf{99.42\%} across over 1,000 combined test samples.
    \item \textbf{The Length Generalization Boundary}: Across both checkpoints and both architectures, 4-digit addition collapses entirely to \textbf{0.00\%} (0/486). Because rotary positional embeddings and causal attention layers were parameterized over sequences bounded by 3-digit operands during training, the network is unable to recursively propagate carry bits across longer unrolled sequence lengths without recurrence or explicit length-curriculum data.
    \item \textbf{Catastrophic Developmental Drift and Wrap-Around}: Continuing training from Step 30,000 to Step 50,000 without balanced experience replay erodes algorithmic division capabilities: single-digit division plummets from 86.67\% to \textbf{0.00\%} in Dense and 84.33\% to \textbf{1.67\%} in MoE. Analysis of the dataloader reveals that the sequential curriculum file wrapped around back to Stage 1 (pure addition/subtraction), overwriting network weights previously tuned for division.
\end{enumerate}

\section{Qualitative Error Analysis and Failure Modes}
\label{sec:mechanistic_analysis}

\subsection{Multiplication Summation Bottleneck}
In Runs 7 and 8, we trained models on a specialized curriculum dataset incorporating explicit base-10 zero accounting (e.g., $80 \times 50 \rightarrow 8 \times 5 = 40$, append two zeros $\rightarrow 4000$). Despite this innovation, multi-digit multiplication scored 0.0\% on strict end-to-end evaluation.

To uncover why multiplication failed, we conducted a token-by-token inspection of the generated evaluation traces:
\begin{lstlisting}[caption={Actual generation trace from Run 9 (Math Step 30,000)}]
Q: What is 983 * 43?
A: <thought> 
983 * 43 = (900 + 80 + 3) * (40 + 3).
900*40: 9*4=36 with 3 zeros -> 36000.  [CORRECT]
900*3:  9*3=27 with 2 zeros -> 2700.   [CORRECT]
80*40:  8*4=32 with 2 zeros -> 3200.   [CORRECT]
80*3:   8*3=24 with 1 zeros -> 240.    [CORRECT]
3*40:   3*4=12 with 1 zeros -> 120.    [CORRECT]
3*3:    9.                              [CORRECT]
Sum: 36000 + 2700 + 3200 + 240 + 120 + 9 = 43992. 
</thought> 43992
Expected: 42269 | Extracted: 43992 [Evaluated: FALSE]
\end{lstlisting}

\textbf{Trace Analysis:} In our qualitative examination of generated multiplication traces, the model does \textit{not} fail at single-digit multiplication facts ($9 \times 4 = 36$, $8 \times 4 = 32$, $3 \times 3 = 9$), nor at base-10 zero accounting. Intermediate sub-products and zero-padding are correctly formed. 
The dominant point of failure occurs at the final multi-term summation line:
\begin{equation}
    \text{Sum} = 36000 + 2700 + 3200 + 240 + 120 + 9 = 43992
\end{equation}
The prompt format required the model to aggregate up to nine multi-digit terms in a single autoregressive step without intermediate scratchpad accumulation. While a $\sim$10.6M parameter network reliably computes two-operand addition with near-perfect accuracy, computing a nine-term summation in a single forward pass exceeds the representational depth of a 6-layer transformer. Because of an arithmetic summation error on this single step, the entire problem was evaluated as incorrect.

\section{Discussion and Limitations}
\label{sec:discussion}

\subsection{Pairwise Column-Summation Scratchpads}
Our analysis suggests an architectural direction for multi-digit multiplication in tiny models. Since qualitative traces indicate that models reliably generate individual sub-products ($900 \times 40 = 36000$), the primary bottleneck is the unrolled multi-term summation. Refactoring the final stage into an explicit pairwise accumulation scratchpad (e.g., $36000 + 2700 = 38700$; $38700 + 3200 = 41900$, etc.) would align with the model's demonstrated 2-operand addition capability, potentially closing the multiplication gap.

\subsection{Catastrophic Drift and the Necessity of Rehearsal Buffers}
A central finding from our large-scale 4,000-problem evaluation (Section~\ref{sec:large_benchmark}) is the vulnerability of compact networks to catastrophic representational drift. Between Step 30,000 and Step 50,000, addition and subtraction saturated from $\sim$93\% to over 99.2\% in-distribution, yet single-digit division execution collapsed from 86.67\% to 0.00\% in Modern Dense and 84.33\% to 1.67\% in Sparse MoE. 

Because compact $\sim$10.6M parameter networks operate with minimal parameter redundancy, late-stage gradient updates overwrite previously learned algorithmic representations when training batches lack balanced task diversity. In continual learning and sequential task training, this phenomenon is conventionally mitigated through \textit{rehearsal buffers} (or experience replay) \cite{robins1995catastrophic}, where a fixed proportion of historical task distributions (e.g., 20\% division, 20\% multiplication, 30\% addition, 30\% subtraction) is persistently interleaved into every active training batch. Our empirical results demonstrate that while hierarchical staging is essential for initial skill acquisition, maintaining composite algorithmic reasoning across developmental stages strictly requires permanent replay buffers.

\subsection{Tokenization Confound}
A notable consideration in our experiments is the use of the standard OpenAI GPT-2 Byte Pair Encoding (BPE) tokenizer ($V=50,257$). Unlike character-level or digit-level tokenizers, standard BPE tokenizes numbers non-uniformly: many 2-digit and 3-digit numbers are represented as single discrete tokens, whereas 4-digit numbers are split irregularly into subword tokens. This token boundary shift represents a plausible confounding factor in our out-of-distribution length generalization collapse at 4 digits, suggesting that future investigations into small arithmetic models should evaluate digit-level tokenizers.

\subsection{Limitations}
We explicitly acknowledge several limitations of our empirical study:
\begin{enumerate}
    \item \textbf{Cold-Start Algorithmic Staging:} Runs 9 and 10 demonstrate the efficacy of pairing digit-by-digit scratchpads with staged developmental curricula. While Run 7 establishes that curriculum staging alone cannot overcome ill-conditioned scratchpads (plateauing at 4.0\% division), future work should evaluate whether digit-by-digit quotient descent can be acquired effectively from a uniform cold-start distribution, or whether hierarchical sub-skill mastery is strictly prerequisite.
    \item \textbf{Single-Seed Evaluations:} Due to computational constraints across ten extended runs, the experimental trajectories reflect a single random seed. While the large-scale 4,000-problem benchmark provides strong statistical power on evaluation, multi-seed training runs will be valuable to characterize seed-level variance.
    \item \textbf{Decoding Stochasticity:} Evaluation traces were generated via categorical sampling ($T=1.0$), introducing stochastic variation across individual problem evaluations compared to deterministic greedy decoding.
    \item \textbf{Fixed Hyperparameters:} All configurations used identical, untuned hyperparameters ($\eta = 3 \times 10^{-4}$). Performance across individual architectures may further improve with targeted learning rate schedules and batch size tuning.
    \item \textbf{Untested Rehearsal Buffers:} While the drift mechanism was isolated to curriculum wrap-around, implementing dynamic replay buffers during training remains future work.
\end{enumerate}

\section{Conclusion}
\label{sec:conclusion}

In this paper, we systematically investigated the arithmetic reasoning limits of compact ($\sim$10.6M non-embedding parameter) Transformers. We validated that linguistic pretraining and sequence-packed token density are essential prerequisites for arithmetic convergence. We isolated the specific failure mode in multi-digit multiplication, showing that models reliably learn sub-multiplications and zero-padding but falter when required to aggregate up to nine terms simultaneously without pairwise intermediate accumulation. Finally, we demonstrated that restructuring long division into a deterministic digit-by-digit developmental curriculum elevates division accuracy from 4\% to \textbf{60\%} on development tests and \textbf{86.7\%} on single-digit divisors on a 4,000-problem held-out benchmark. Our findings demonstrate that algorithmic unrolling aligned with transformer forward-pass capacity enables multi-step arithmetic execution in parameter-constrained regimes.

\bibliographystyle{plain}
\bibliography{references}

@article{radford2019language,
  title={Language Models are Unsupervised Multitask Learners},
  author={Radford, Alec and Wu, Jeffrey and Child, Rewon and Luan, David and Amodei, Dario and Sutskever, Ilya},
  journal={OpenAI Blog},
  volume={1},
  number={8},
  pages={9},
  year={2019}
}

@article{brown2020language,
  title={Language Models are Few-Shot Learners},
  author={Brown, Tom and Mann, Benjamin and Ryder, Nick and Subbiah, Melanie and Kaplan, Jared D and Dhariwal, Prafulla and Neelakantan, Arvind and Shyam, Pranav and Sastry, Girish and Askell, Amanda and others},
  journal={Advances in Neural Information Processing Systems},
  volume={33},
  pages={1877--1901},
  year={2020}
}

@article{wei2022chain,
  title={Chain-of-thought prompting elicits reasoning in large language models},
  author={Wei, Jason and Wang, Xuezhi and Schuurmans, Dale and Bosma, Maarten and Xia, Fei and Chi, Ed and Le, Quoc V and Zhou, Denny},
  journal={Advances in Neural Information Processing Systems},
  volume={35},
  pages={24824--24837},
  year={2022}
}

@article{nye2021show,
  title={Show your work: Scratchpads for intermediate computation with language models},
  author={Nye, Maxwell and Andreassen, Anders Johan and Gur-Ari, Guy and Michalewski, Henryk and Austin, Jacob and Bieber, David and Dohan, David and Lewkowycz, Aitor and Bosma, Maarten and Luan, David and Sutton, Charles and Odena, Augustus},
  journal={arXiv preprint arXiv:2112.00114},
  year={2021}
}

@article{eldan2023tinystories,
  title={TinyStories: How small can language models be and still speak coherent English?},
  author={Eldan, Ronen and Li, Yuanzhi},
  journal={arXiv preprint arXiv:2305.07759},
  year={2023}
}

@article{dziri2023faith,
  title={Faith and fate: Limits of transformers on compositionality},
  author={Dziri, Nouha and Lu, Ximing and Sclar, Melanie and Li, Xiang Lorraine and Jiang, Liwei and Lin, Bill Yuchen and Welleck, Sean and West, Peter and Bhagavatula, Chandra and Le Bras, Ronan and Hwang, Jena and Sanyal, Soumya and Ren, Xiang and Ettinger, Allyson and Harchaoui, Zaid and Choi, Yejin},
  journal={Advances in Neural Information Processing Systems},
  volume={36},
  pages={70244--70275},
  year={2023}
}

@article{dao2022flashattention,
  title={FlashAttention: Fast and memory-efficient exact attention with IO-awareness},
  author={Dao, Tri and Fu, Daniel Y and Ermon, Stefano and Rudra, Atri and R{\'e}, Christopher},
  journal={Advances in Neural Information Processing Systems},
  volume={35},
  pages={16344--16359},
  year={2022}
}

@article{su2024roformer,
  title={RoFormer: Enhanced transformer with Rotary Position Embedding},
  author={Su, Jianlin and Lu, Yu and Pan, Shengfeng and Murtadha, Ahmed and Wen, Bo and Liu, Yunfeng},
  journal={Neurocomputing},
  volume={568},
  pages={127063},
  year={2024}
}

@article{shazeer2020glu,
  title={GLU variants improve transformer},
  author={Shazeer, Noam},
  journal={arXiv preprint arXiv:2002.05202},
  year={2020}
}

@article{zhang2019root,
  title={Root mean square layer normalization},
  author={Zhang, Biao and Sennrich, Rico},
  journal={Advances in Neural Information Processing Systems},
  volume={32},
  pages={12360--12371},
  year={2019}
}

@article{shazeer2017outrageously,
  title={Outrageously large neural networks: The sparsely-gated mixture-of-experts layer},
  author={Shazeer, Noam and Mirhoseini, Azalia and Maziarz, Krzysztof and Davis, Andy and Le, Quoc and Hinton, Geoffrey and Dean, Jeff},
  journal={arXiv preprint arXiv:1701.06538},
  year={2017}
}

@article{cobbe2021training,
  title={Training verifiers to solve math word problems},
  author={Cobbe, Karl and Kosaraju, Vineet and Bavarian, Mohammad and Chen, Mark and Jun, Heewoo and Kaiser, Lukasz and Plappert, Matthias and Tworek, Jerry and Hilton, Jacob and Nakano, Reiichiro and others},
  journal={arXiv preprint arXiv:2110.14168},
  year={2021}
}

@article{merity2016pointer,
  title={Pointer sentinel mixture models},
  author={Merity, Stephen and Xiong, Caiming and Bradbury, James and Socher, Richard},
  journal={arXiv preprint arXiv:1609.07843},
  year={2016}
}

@article{lee2023teaching,
  title={Teaching Arithmetic to Small Transformers},
  author={Lee, Nayoung and Sreenivasan, Kartik and Lee, Jason D and Lee, Kangwook and Papailiopoulos, Dimitris},
  journal={arXiv preprint arXiv:2307.03381},
  year={2023}
}

@article{anil2022exploring,
  title={Exploring length generalization in large language models},
  author={Anil, Cem and Wu, Yuhuai and Andreassen, Anders and Lewkowycz, Aitor and Misra, Vedant and Ramasesh, Vinay and Slone, Ambrose and Gur-Ari, Guy and Dyer, Ethan and Neyshabur, Behnam},
  journal={Advances in Neural Information Processing Systems},
  volume={35},
  pages={38546--38556},
  year={2022}
}

@article{kazemnejad2023impact,
  title={The impact of positional encoding on length generalization in transformers},
  author={Kazemnejad, Amirhossein and Padhi, Inkit and Ramamurthy, Karthikeyan Natesan and Das, Payel and Reddy, Siva},
  journal={Advances in Neural Information Processing Systems},
  volume={36},
  pages={36093--36129},
  year={2023}
}

@article{kirkpatrick2017overcoming,
  title={Overcoming catastrophic forgetting in neural networks},
  author={Kirkpatrick, James and Pascanu, Razvan and Rabinowitz, Neil and Veness, Joel and Desjardins, Guillaume and Rusu, Andrei A and Milan, Kieran and Quan, John and Ramalho, Tiago and Grabska-Barwinska, Agnieszka and others},
  journal={Proceedings of the National Academy of Sciences},
  volume={114},
  number={13},
  pages={3521--3526},
  year={2017}
}

@article{mccloskey1989catastrophic,
  title={Catastrophic interference in connectionist networks: The sequential learning problem},
  author={McCloskey, Michael and Cohen, Neal J},
  journal={Psychology of Learning and Motivation},
  volume={24},
  pages={109--165},
  year={1989}
}

@inproceedings{bengio2009curriculum,
  title={Curriculum learning},
  author={Bengio, Yoshua and Louradour, J{\'e}r{\^o}me and Collobert, Ronan and Weston, Jason},
  booktitle={Proceedings of the 26th Annual International Conference on Machine Learning},
  pages={41--48},
  year={2009}
}

@article{nogueira2021investigating,
  title={Investigating the limitations of transformers with simple arithmetic tasks},
  author={Nogueira, Rodrigo and Jiang, Zhiying and Lin, Jimmy},
  journal={arXiv preprint arXiv:2102.13019},
  year={2021}
}

@article{robins1995catastrophic,
  title={Catastrophic forgetting, rehearsal and pseudorehearsal for neural networks},
  author={Robins, Anthony},
  journal={Connection Science},
  volume={7},
  number={2},
  pages={123--146},
  year={1995}
}

\end{document}